\documentclass[runningheads]{llncs}
\usepackage{xcolor}
\usepackage{graphicx}
\begin{document}
\title{RRAML: Reinforced Retrieval Augmented Machine Learning} 
%
%
\author{Anonymous Authors}
\author{Andrea Bacciu\inst{1} \and Florin Cuconasu\inst{1} \and Federico Siciliano\inst{1} \and Fabrizio Silvestri\inst{1} \and Nicola Tonellotto\inst{2} \and Giovanni Trappolini\inst{1}\\[2ex]
\{surname\}@diag.uniroma1.it\inst{1}, nicola.tonellotto@unipi.it\inst{2}}
\authorrunning{Bacciu, Cuconasu, Siciliano, Silvestri, Tonellotto, Trappolini, 2023}
\institute{
  Sapienza University of Rome \and 
  University of Pisa  
}
\institute{Sapienza University of Rome \and University of Pisa}

%
%
\maketitle              
\begin{abstract}
The emergence of large language models (LLMs) has revolutionized machine learning and related fields, showcasing remarkable abilities in comprehending, generating, and manipulating human language.
However, their conventional usage through API-based text prompt submissions imposes certain limitations in terms of context constraints and external source availability.
LLMs suffer from the problem of hallucinating text, and in the last year, several approaches have been devised to overcome this issue: adding an external Knowledge Base or an external memory consisting of embeddings stored and retrieved by vector databases. In all the current approaches, though, the main issues are: (i) they need to access an embedding model and then adapt it to the task they have to solve; (ii) in case they have to optimize the embedding model, they need to have access to the parameters of the LLM, which in many cases are ``black boxes''.
To address these challenges, we propose a novel framework called Reinforced Retrieval Augmented Machine Learning (RRAML).
RRAML integrates the reasoning capabilities of LLMs with supporting information retrieved by a purpose-built retriever from a vast user-provided database.
By leveraging recent advancements in reinforcement learning, our method effectively addresses several critical challenges. Firstly, it circumvents the need for accessing LLM gradients. Secondly, our method alleviates the burden of retraining LLMs for specific tasks, as it is often impractical or impossible due to restricted access to the model and the computational intensity involved.
Additionally, we seamlessly link the retriever's task with the reasoner, mitigating hallucinations and reducing irrelevant and potentially damaging retrieved documents.
We believe that the research agenda outlined in this paper has the potential to profoundly impact the field of AI, democratizing access to and utilization of LLMs for a wide range of entities.

\end{abstract}

\section{Introduction}

The advent of Large Language Models (LLMs) has brought about a paradigm shift in machine learning and its related disciplines.
LLMs~\cite{brown2020language,scao2022bloom,openai2022chatgpt,touvron2023llama,bacciu2023fauno} have exhibited unprecedented capabilities in understanding, generating, and manipulating the human language.
Famously, ChatGPT \cite{openai2022chatgpt} has entered the public space by reaching one million users in a matter of days.
The way these models are usually used is through API that only allows submitting a textual prompt and getting back from the server the generated text. 
However, this causes an immediate limitation: all information must be passed through this context, and we know transformer-based models do not scale nicely.
Even if they did, API costs are charged on the basis of their usage. Therefore, using long contexts would be expensive.
Even if one had the resources to run their own LLM, the costs of training and of the hardware infrastructure, and the environmental impact should be considered.
There is an impendent need, though, to accommodate the enormous power of those models to specific user needs by making sure that they could use the reasoning capabilities of LLMs, through in-context learning \cite{brown2020language} on their data.\\
A solution is to adopt a retrieval-augmented approach \cite{lewis2020retrieval,xie2023factual}.
In this setting, a retriever is used to filter out relevant information to be passed as context to the reasoner.
This generates a new problem, however, namely that the retriever and the reasoner are not aligned \cite{trappolini2023multimodal,thorne2021database,thorne2021natural}.
In particular, the retriever might not be trained on the task of interest to the user.
Moreover, the retriever might actually provide ``dangerous” pieces of information to the reasoner, as proved in \cite{artsiomdang}, leading to poor results and, more importantly, to hallucinations.

Ideally, one would have to fine-tune these models to account for these issues.
Within this setting, fine-tuning the model for a given task is technically impossible.
We asked ourselves: ``\textit{Is it still possible to use the API that gatekeeps those powerful LLMs on our data without the need for fine-tuning?}''
We show that this question has a positive answer and in this paper, we propose a novel framework, Reinforced Retrieval Augmented Machine Learning (RRAML), in which we combine the reasoning capabilities of large foundational models enhanced by the provision of supporting relevant information provided by a retriever that searches them in a large database.
In this setting, an efficient retriever model is tasked to search for relevant information in an arbitrarily large database of data provided by users. Once this set of relevant data has been retrieved, it is forwarded to the reasoner (a large foundational model such as ChatGPT, for instance) through its API to ``reason" on the input and produce an adequate result.
In particular, we plan to overcome current limitations, namely that the retriever's task is detached from that of the reasoner, reducing in such a way the tendency of LLM to hallucinate and diminishing the number of damaging documents (as defined in \cite{carmel2022ir,sauchuk2022role,trappolini2023multimodal}) returned by the retriever.
The approach we devise in this research work exploits recent advances in reinforcement learning.
Recently, in fact, reinforcement learning techniques like PPO \cite{schulman2017proximal} have been used to improve large foundational models with human feedback where the loss is non-differentiable.
We propose to link the training phase of the retriever to the final task outcome by the use of a purposefully crafted reward model that depends either on human feedback or on the specific characteristics of the task data.
The RL technique also offers the advantage of not requiring fine-tuning an LLM as a reasoner, which can be considered a black box in this setting, and exchanged freely.

Finally, we argue that the research agenda we lay out in this paper has the potential to hugely impact the field of AI and democratize the access and use of these large foundational models to a large set of entities.

\section{Methodology}
\begin{figure}[!ht]
\centering
\includegraphics[width=\textwidth]{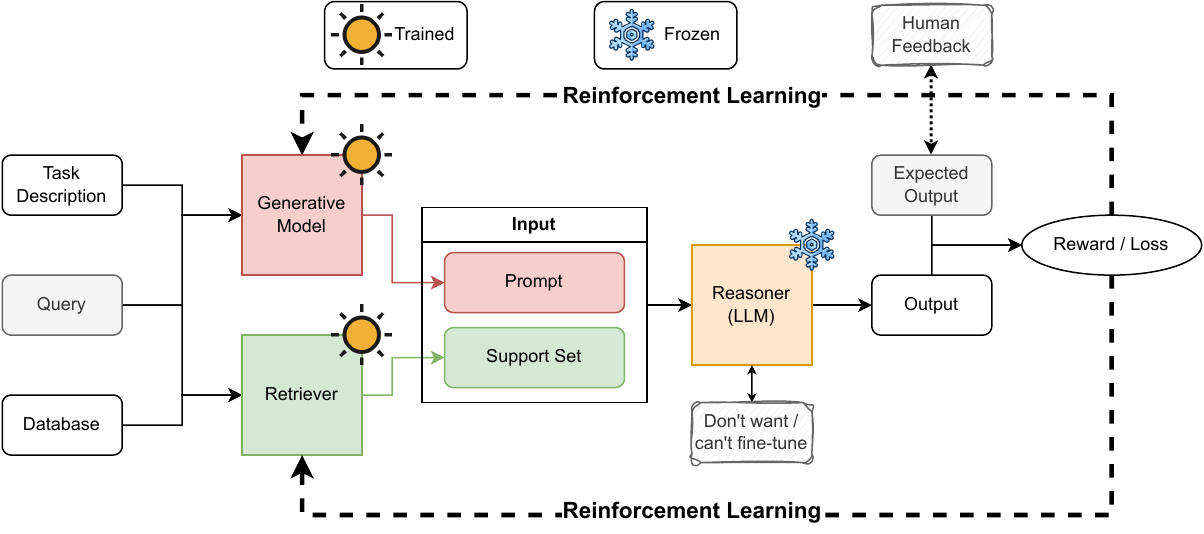}
\caption{High-level design of the RRAML framework. On the left side, there are the three inputs: Task Description, user's query, and a database that represents the external knowledge used to augment/update the reasoner. Then, we present the overall architecture flow with the Retriever, Generative Language Model, and Reasoner. Finally, how the reward is computed and propagated in the Generative Language Model and Retriever.}
\label{fig:design}
\end{figure}

The system takes as input a task description, a query, and a database and gives as output the response generated by a reasoner.
The overall system architecture, shown in Figure \ref{fig:design}, consists of three main components: a Generative Language Model, a Retriever, and a Reasoner (typically an LLM).

More in detail, the Generative Language Model takes the \textit{task description} and \textit{query} as input and generates a prompt. The Retriever takes the \textit{query} and the \textit{database} as input and outputs a support set, which is then concatenated with the \textit{query} and passed to the Reasoner.

\subsection{Data}
The data is a critical component of the framework: the \textit{task description} guides the generation of an appropriate prompt, the \textit{query} represents the user request, and the \textit{database} provides the data needed by the reasoner to perform the task.

\paragraph{Task Description}
The \textit{task description} is a string that defines the nature of the task, possibly with expected results, that the user wants to perform. 
For example, if the user wants to generate a summarization of multiple news articles, a possible \textit{task description} could be ``News Summarization". If the user wants to perform question answering on a vast document collection, the \textit{task description} could be ``Question Answering".


\paragraph{Query}
The \textit{query} represents the user's need. The Retriever will operate on the \textit{database} w.r.t to the user's query, and the resulting data is input for the task. For example, if the user wants to summarize a collection of news articles, the \textit{query} could be the topic the user is interested in.
If the user wants to answer a specific \textit{query}, this becomes the actual question.

\paragraph{Database}
The \textit{database} is a collection of public or private data (or documents) that can be queried to provide relevant information to satisfy the user's information needs.
The database represents the knowledge needed by the Reasoner to perform the task.
The data stored in the \textit{database} will depend on the specific task and may include text, images, audio, and other data types (as in \cite{trappolini2023multimodal}). 
For example, if the user wants to summarize multiple news articles, the \textit{database} could be an indexed collection of articles.
If the user wants to perform Question Answering, the \textit{database} may consist of facts related to a particular topic (as in \cite{thorne2021database,thorne2021natural}).


\subsection{Models}

\paragraph{Generative Language Model}
The Generative Language Model component of the framework is responsible for generating textual instructions based on the input \textit{Task Description} and \textit{Query} that maximize the rewards w.r.t Reasoner.
Specifically, it receives a string representing the task to be performed (\textit{Task Description}) and a query (\textit{Query}) that represents the user's request.
The Generative Language Model then generates a textual prompt that is relevant to the query and the task by performing automatic prompt engineering.

\paragraph{Retriever}
The Retriever component of the framework is responsible for retrieving relevant data from the Database based on the user's query.
We refer to the Retriever outputs as support set (as in \cite{thorne2021database,thorne2021natural}).
A support set is a subset of the data from the Database that either directly answers the given query or contributes to the final answer.

%




\paragraph{Prompt Aggregator}
This component is responsible for processing the input required by the \textit{Reasoner}. 
In its simplest form, it just needs to concatenate the prompt generated by the Generative Language Model with the Support Set provided by the Retriever. 
However, in a more complex version, it may need to rework the prompt based on the number of support sets received to ensure that the LLM can provide a coherent response.
For example, if the Retriever provides two support sets, the Prompt Aggregator may need to split the prompt into two parts and concatenate each part with one of the support sets.

\paragraph{Reasoner}
The Reasoner is responsible for generating the answer to the user's query based on the final prompt generated by the Prompt Aggregator.
The Reasoner can be a pre-trained model like GPT or a custom-trained model specific to the task at hand.
The output of the LLM is a textual response, which can be further parsed to comply with the intended output.

\subsection{Reinforcement Learning}
The Reinforcement Learning (RL) part of the framework is responsible for fine-tuning the Generative Language Model (GLM) and Retriever based on the computed reward.
The RL is a crucial part of RRAML, it will be used to constantly improve the GLM and Retriever.
As mentioned earlier, the retriever will get a penalty if some of his recommendations will leads the Reasoner to a hallucinate, for example by adding damaging documents.
The RL allows use to integrate and augment the signals in the training of these models, going beyond the data present in their training set, ensuring that they are aligned with the environment (i.e., the reasoner and the final task).

\paragraph{Reward}
The reward function can be defined based on the similarity between the generated output and the expected output and it can be estimated by training a Reward Model \cite{schulman2017proximal}.

\paragraph{RL algorithm}
The specific RL method which can be used is Deep Q-Networks (DQN) \cite{mnih2015human}, which is a model-free RL algorithm that learns to maximize the cumulative reward over time.
DQN combines Q-Learning, which is a RL algorithm that learns the optimal action-value function, with a Deep Neural Network to approximate the action-value function.
In the proposed framework, DQN is used to train the Generative Language Model and the Retriever to maximize the reward obtained from user feedback.
The update process is performed by backpropagating the reward signal through the neural networks using Stochastic Gradient Descent (SGD). The weights of the neural networks are updated in the direction that maximizes the expected reward, using the Q-Learning update rule.
The update is performed iteratively until convergence, which is achieved when the expected reward stops improving.

\paragraph{Human-in-the-loop}
Human preferences can be incorporated into our ML system by allowing users to provide feedback on the system's output.
This feedback will be used to compute the reward for the RL algorithm and will help improve the performance of the overall system over time.
We acknowledge that some tasks may not have a clear expected output or may require additional context that is not available in the input data.
In these cases, we will leverage human-in-the-loop approaches to provide additional context and guidance to the system.
For example, crowd-sourcing platforms or internal subject matter experts can be used to provide feedback on the system's output and help train the model on more complex tasks.


\section{Use Case Example}
RRAML proves to be effective in many applications. Consider a situation where a company possesses a private database, which consists of factual information expressed in natural language, and they need to apply reasoning to this data. The volume of their data may exceed the context capacity of the LLM, and fine-tuning is not an option, for pricing/environmental impact or because the LLM is served by other company APIs.
To tackle this challenge, RRAML uses its retriever to get only the relevant facts within the context, enabling the LLM to reason over them.

For instance, suppose a company has an employee list, projects that employees are currently or were previously assigned to, and performance evaluation grids with text-based feedback from superiors. The company might want to assign employees to a new project on a specific topic. To do so, it is necessary to input the information contained in these data to the LLM. However, due to capacity constraints, the entire data cannot fit within the context. Therefore, the retriever has to return a subset of this information, perhaps excluding data on projects from the distant past, employees who are already overburdened with multiple projects, or employees who have never worked on a project related to the same topic.

\section{Related Work}

Recent years have seen the emergence of large language models. Starting from the first Generative Pre/Training Model, better known as GPT \cite{radford2018improving}, these kinds of large language models have rapidly improved.
GPT-4 \cite{openai2023gpt4} is the most recent iteration, but in the meanwhile, many have rushed to propose their own version.
Google has recently released BARD\footnote{\url{https://bard.google.com/}}, while Meta has proposed their own take on LLM with LLaMA \cite{touvron2023llama}.
The research community has also capitalized its effort by releasing several open source LLM of different sizes, like Bloom \cite{scao2022bloom}, Dolly\footnote{\url{https://github.com/databrickslabs/dolly}}, and RWKV \cite{PENG_RWKV-LM_2021}.
However, all these models fail to scale to a larger context size, either by excessive computational costs or by ``losing it in the middle", as shown in \cite{liu2023lost}.
\\
To address this context-length limitation, some have tried to incorporate external knowledge into LLMs \cite{ghazvininejad2018knowledge,dinan2018wizard,peng2023check}.
In particular, in ``Retrieval-enhanced machine learning" \cite{zamani2022retrieval}, authors have envisioned a framework in which retrieval systems can enhance the performance of a machine learning model. 
More recently, there have been attempts of jointly training retrieval models with LLMs \cite{lewis2020retrieval,zhang2022retgen}, notably, the line of research on neural databases, in which the authors tried to replace a traditional database with a neural framework removing the need for a schema \cite{thorne2021natural,thorne2021database,trappolini2023multimodal}.
However, all these works assume full access to the reasoner module, which is not the case for most users in practice.
\\
To overcome this limitation, many have tried to craft systems that are able to deliver an optimized prompt that is input to the LLM.
For instance, the research conducted by \cite{lu2021fantastically} demonstrated a substantial influence of the sequence in which prompts are presented on the ultimate performance of the task. Meanwhile, a study by Nie et al. \cite{nie2022improving} highlighted that the performance is susceptible to the arrangement of the examples in the prompt, prompt templates, and the in-context instances in the prompt.
Lester et al. \cite{lester2021power} suggested a method to enhance task performance by adding adjustable tokens during fine-tuning. 
LLM-AUGMENTER iteratively revises \cite{peng2023check} to improve the model response.
\\
All the works introduced above do not improve on the retriever, which is assumed fixed.
In our work, we propose to finetune the retriever in conjunction with the reasoner to improve on results.
Since the feedback is non-differentiable we resort to reinforcement learning.
In particular, recent formulation such as Proximal Policy Optimization (PPO) \cite{engstrom2020implementation} make use of a differentiable neural reward module to include and account for generally non-differentiable feedback, like in the case of reinforcement learning with human feedback (RLHF).


\section{Conclusions}
In conclusion, RRAML provides a promising framework for building intelligent interfaces to interact with large language models like GPT. By combining a generative language model with a retriever, this approach can effectively improve the performance of language models and help them understand user intents better.

However, this approach also comes with several challenges and uncertainties, such as the need for a large amount of training data, the potential for bias in the data and models, and the difficulty of balancing the trade-offs between generative and retrieval-based approaches.

Despite these challenges, RRAML holds great promise for creating more intelligent, natural, and effective interfaces for interacting with language models. We hope that this paper has provided a useful overview of this approach and its potential applications, and we look forward to further research and development in this exciting area.

\clearpage

\bibliographystyle{splncs04}
\bibliography{bib}
%




\end{document}